\documentclass{article}

\newcommand{\enum}[1]{\label{eq:#1}}

\newcommand{\eref}[1]{(\ref {eq:#1})}

\usepackage{amsmath, amsthm, amssymb,subfloat}
\usepackage{tikz,graphics,color,float,epsf,caption,subcaption}
\usepackage{algorithm}
\usepackage[noend]{algpseudocode}
\usepackage{makecell}

\newtheorem{defn}{Defenition}

\newcommand{\be}{\begin{equation}}
\newcommand{\ee}{\end{equation}}

\newcommand{\bear}{\begin{eqnarray}}
\newcommand{\eear}{\end{eqnarray}}

\usepackage[english]{babel}
\usepackage[utf8x]{inputenc}
\usepackage[T1]{fontenc}

\usepackage{amsmath}
\usepackage{graphicx}
\usepackage[colorinlistoftodos]{todonotes}
\usepackage[colorlinks=true, allcolors=blue]{hyperref}

\usepackage[T1]{fontenc}    
\usepackage{hyperref}       
\usepackage{url}            
\usepackage{booktabs}       
\usepackage{amsfonts}       
\usepackage{nicefrac}       
\usepackage{microtype}      

\title{A general metric for identifying adversarial images}

\author{
  Siddharth Krishna Kumar \\
  Upwork Inc.\\
  441 Logue Avenue,\\
  Mountain View, CA 94043 \\
  \texttt{siddharthkumar@upwork.com} \\
}

\begin{document}

\maketitle

\begin{abstract}

It is well known that a determined adversary can fool a neural network by making imperceptible adversarial perturbations to an image. Recent studies have shown that these perturbations can be detected even without information about the neural network if the strategy taken by the adversary is known beforehand. Unfortunately, these studies suffer from the generalization limitation -- the detection method has to be recalibrated every time the adversary changes his strategy. In this study, we attempt to overcome the generalization limitation by deriving a metric which reliably identifies adversarial images even when the approach taken by the adversary is unknown. Our metric leverages key differences between the spectra of clean and adversarial images when an image is treated as a matrix.  Our metric is able to detect adversarial images across different datasets and attack strategies without any additional re-calibration. In addition, our approach provides  geometric insights into several unanswered questions about adversarial perturbations.

\end{abstract}

\section{Introduction}

In recent years, neural networks have been shown to be a powerful and versatile tool for image detection. Unfortunately, these networks are not perfect; multiple studies have shown that the output of these networks can be dramatically altered by making imperceptible {\it{adversarial perturbations}} to a correctly classified image. These adversarial perturbations can be generated using many  {\it{attack strategies}} (\cite{moosavi2016deepfool},\cite{goodfellow2014explaining},\cite{madry2017towards} to state a few), and can have differing {\it{strengths}}. These adversarial perturbations pose a serious security risk \cite{mcdaniel2016machine},  and could jeopardize several of the applications that rely on this technology like self driving cars
\cite{evtimov2017robust}.


The ruling hypothesis about adversarial perturbations is that they induce critical changes in the geometry of the clean images that are perceived by neural networks, but not by humans. Strong evidence for this hypothesis comes from the work of \cite{grosse2017statistical} and \cite{gong2017adversarial}. These papers show that
if the attack strategy and the maximum strength of the perturbation is known beforehand, then we can identify the adversarial images in the dataset even without information about the neural network. Unfortunately, the methods used in these papers do not `generalize' -- they need to be re-calibrated every time the adversary changes his strategy for generating adversarial images.

This raises the question, are there differences between clean and adversarial images that are independent of the attack strategy and its strength ? In this paper, we show that the answer appears to be yes, by deriving a metric which separates clean and adversarial images with high probability across a variety of datasets and attack strategies. To derive our metric, we treat an image as an `image matrix', and identify critical geometrical differences between the spectra of clean and adversarial image matrices that are independent of the approach used to generate the adversarial images. Our metric is `universal' in the sense that once setup, it does not need to be tuned for different attack strategies and perturbation strengths. Our approach provides numerous insights into the inner workings of adversarial perturbations and provides a possible answer for an open question posed in \cite{dziugaite2016study}.

\section{Adversarial Images}

In this section, we provide a brief  overview of our current knowledge on adversarial examples.

\subsection{Adversarial images and defenses against them}

Adversarial images were first investigated by \cite{szegedy2013intriguing}, where the authors observed that by applying a small carefully crafted perturbation to an image, a neural network can be made to misclassify the image with high confidence. Ever since, there has been an explosion in methods to create adversarial images (\cite{moosavi2016deepfool},\cite{goodfellow2014explaining},\cite{carlini2017towards} to state a few), with every new method being more sophisticated than the previous one. An overview of many of the attack methods can be found in \cite{akhtar2018threat}. 

Given the great harm these adversarial perturbations can cause, there has been a strong recent focus on {\it{defenses}} to nullify the effects of these adversarial perturbations. The defenses can be broadly classified into two groups - the model specific defenses and the model agnostic defenses. The model specific defenses attempt to tweak the properties of the neural network to make them more robust to adversarial perturbations. These defenses include methods like adversarial retraining \cite{kurakin2016adversarial}, and defensive distillation \cite{papernot2016distillation} among others. The model agnostic strategies attempt to
 make the neural network robust to adversarial perturbations by making tweaks to the input image which nullify the effect of the adversarial images. This defenses include methods like quilting \cite{guo2017countering} and image compression \cite{dziugaite2016study} among others. 
 
Although we have made considerable progress in making neural networks robust to adversarial perturbations, the problem is far from being resolved; \cite{carlini2017adversarial} show how a determined adversary can outwit some of our best known defenses available to date. 



\begin{figure}

\centering
\resizebox{0.6\textwidth}{!}{
    \begin{subfigure}{.3\textwidth}
        \fbox{\includegraphics[width=\linewidth]{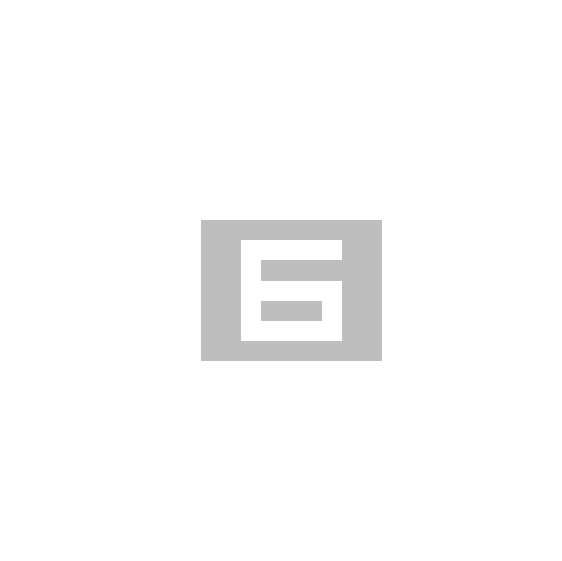}}
        \caption{}\label{fig:fig_1a}
    \end{subfigure} %
     \hspace{1cm}
    \begin{subfigure}{.3\textwidth}
        \fbox{\includegraphics[width=\linewidth]{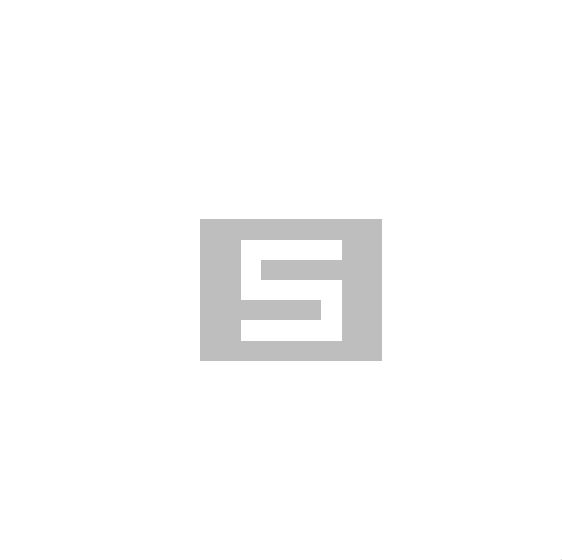}}
        \caption{}\label{fig:fig_1b}
    \end{subfigure} %
}
\caption{\label{fig:fig_1}Changing a single pixel in the image on the left converts it into the image on the right. The overall difference between the two images is less than 1\% in the $L_{2}$ sense.  }
\end{figure}




\subsection{The geometry of adversarial perturbations}

Our limited insights into the geometry of adversarial images comes primarily from the work of Goodfellow and his colleagues  \cite{goodfellow2014explaining},\cite{tramer2017space}. The key findings are that 1) adversarial images lie in continuous regions of high dimensional spaces and 2) that adversarial perturbations are transferrable because a significant fraction of the image subspaces are shared across different model architectures.

In a recent paper, \cite{elsayed2018adversarial} demonstrate that minor perturbations can fool humans too. They provide the example of a hybrid cat-dog image, which when perturbed slightly one way looks more like a cat, and which perturbed slightly the other way looks more like a dog. The question still remains whether minor perturbations of a clean image can dramatically alter human perception. 

The answer to this question is yes, as is demonstrated in Figure \ref{fig:fig_1}. By making a less than 1\% change to figure \ref{fig:fig_1a} in the $L_{2}$ sense, the perception of the image can be changed from the number 6 to the number 5 (figure \ref{fig:fig_1b}). Coincidentally, figures \ref{fig:fig_1a} and \ref{fig:fig_1b} differ by only a single pixel. Thus, analogous to the one-pixel attack in neural networks \cite{su2017one}, it is possible to dramatically alter human perception by changing a single pixel in the image.

\section{The matrix view of images}

Most studies analyzing adversarial perturbations interpret an image, $I$, of dimension $M \times N \times K$ as a long vector of dimension ($MNK$,1). In this paper, we take a different approach, and represent the image as a block-diagonal `image matrix', ${\bf{I}}$, defined as  
\[
  {\bf{I}} =
  \begin{bmatrix}
    {\bf{I}}_{1} & 0 & 0 & \dots  & 0\\
    0 & {\bf{I}}_{2} & 0 & \dots  & 0\\
    \vdots & \vdots & \ddots & \vdots & \vdots \\
    0  & \dots & 0 & {\bf{I}}_{k-1} & 0\\
    0 & 0  & \dots & 0  & {\bf{I}}_{k}
\end{bmatrix}
\],

where ${\bf{I}}_{k}$  is the $M \times N$ matrix representing the $k^{th}$ channel of the image. Let ${\bf{X}}$ and ${\hat{\bf{X}}}$ denote the image matrices for clean and adversarial images respectively. These matrices satisfy the relationship

\be 
{\hat{\bf{X}}} = {\bf{X}} + {\bf{E}},
\ee 

where ${\bf{E}}$ is a block diagonal matrix with small entries denoting the adversarial perturbation. For the remainder of our analysis, ${\bf{X}}$ and ${\hat{\bf{X}}}$ are referred to as the clean image matrix and the adversarial image matrix respectively.

In this section, we will show that there are important geometric differences between ${\bf{X}}$ and 
${\hat{\bf{X}}}$, and that these differences can be used to identify adversarial images with high probability. We begin by providing a brief overview of the results from matrix theory that are relevant to our analysis. More details on the subject can be found in \cite{horn1990matrix} and \cite{kato2013perturbation}.

\begin{figure}
\centering

    \begin{subfigure}{.3\textwidth}
       \includegraphics[width=\linewidth]{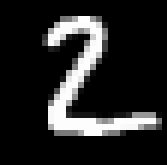}
        \caption{}\label{fig:fig_2a}
    \end{subfigure} %
     \hspace{1cm}
    \begin{subfigure}{.55\textwidth}
        \includegraphics[width=\linewidth]{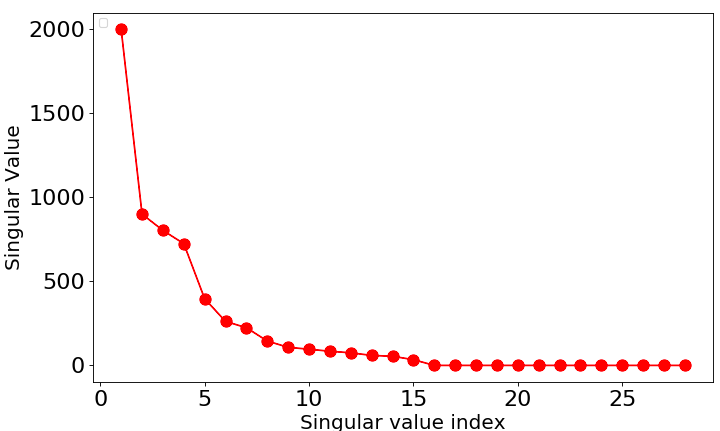}
        \caption{}\label{fig:fig_2b}
    \end{subfigure} %
  \caption{\label{fig:fig_2}{\bf{Left}}: A randomly chosen figure from the MNIST dataset. The image is a $28 \times 28$ grayscale representation of the number 2. {\bf{Right}}: The singular values of the image matrix described on the left. Note that the first 5 singular values are much larger than the remaining. }

\end{figure} 

\subsection{The geometry of a matrix}

With $P = Min(M,N)$, ${\bf{X}}$ can be factored using the Singular Value Decomposition (SVD) as 

\be 
{\bf{X}} = \sum\limits_{i=1}^{i=P} s_{i}u_{i}v_{i}^{T} \enum{svd1}
\ee 

where $s_{i}$, $u_{i}$ and $v_{i}$ are the $i^{th}$ singular value and corresponding left singular vector and right singular vector respectively. In \eref{svd1}, the left and right singular vectors form an orthogonal set, i.e., $u^{T}_{i}u_{j} = 0$ and $v^{T}_{i}v_{j} = 0$ for every $i \neq j$. The singular values and vectors uniquely define ${\bf{X}}$ up to a sign. 

The singular values of a randomly chosen clean image matrix is plotted in figure \ref{fig:fig_2}. From the figure, we note that the first few singular values of ${\bf{X}}$ are much larger than the remaining. These larger singular values contain more information about the geometry of the image than the smaller singular values. The top $k$ singular values of ${\bf{X}}$ contain a fraction $ r = \left(\sum\limits_{i=1}^{i=k} s_{i}^{2}\right)/\left(\sum\limits_{i=1}^{i=P} s_{i}^{2}\right)$  of the total energy contained in an image. Therefore, for sufficiently large values of $k$, a compression of ${\bf{X}}$ defined as ${\bf{X_k}} = \sum\limits_{i=1}^{i=k} s_{i}u_{i}v_{i}^{T}$ will contain most of the information contained in ${\bf{X}}$.




\begin{figure}
\centering
    \begin{subfigure}{.45\textwidth}
       \includegraphics[width=\linewidth]{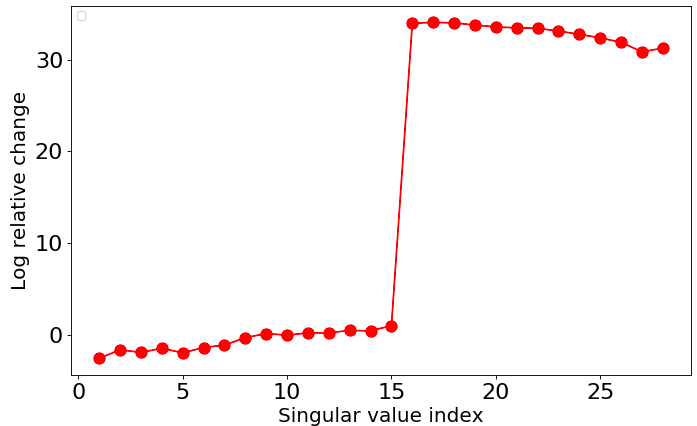}
        \caption{}\label{fig:fig_3a}
    \end{subfigure} %
     \hspace{1cm}
    \begin{subfigure}{.45\textwidth}
        \includegraphics[width=\linewidth]{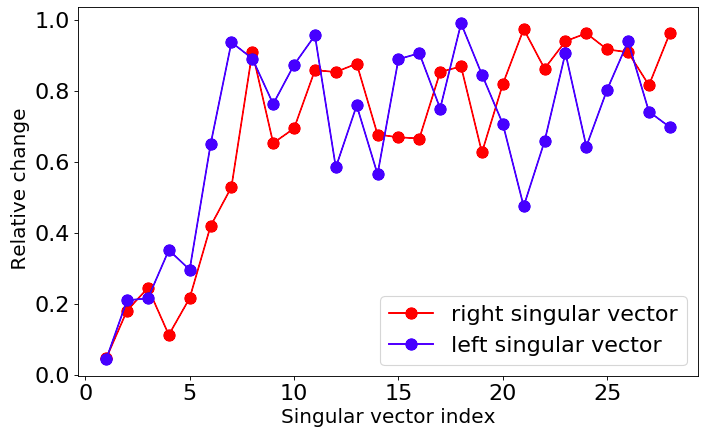}
        \caption{}\label{fig:fig_3b}
    \end{subfigure} %
    \caption{\label{fig:fig_3}Relative change in the singular values (left) and singular vectors (right) of the clean image matrix, when an adversarial perturbation is applied to figure \ref{fig:fig_2a}. The relative change in the small singular values (index 5 and larger according to Figure \ref{fig:fig_2b}) and corresponding singular vectors are large. Note that the y-axis on the figure on the left is in the log scale.}
\end{figure}

\subsubsection{Matrix norms}

The singular values describe a notion of distance in matrices and therefore, not surprisingly, there is a strong connection between matrix norms and singular values; the 2-norm of the image matrix is given by $||{\bf{X}}||_{2} = s_{1} $, and the Frobenius norm of the image matrix is given by $||{\bf{X}}||_{F} = ||Vec({\bf{X}})||_{2} =\sqrt{\sum\limits_{i=1}^{i=P}s_{i}^{2}}$.
Most papers \cite{goodfellow2014explaining},\cite{carlini2017towards},\cite{papernot2016distillation} on adversarial perturbations define the `strength' of an adversarial perturbation in terms of the maximum allowable distortion in an image. This corresponds to the max-norm of ${\bf{X}}$ defined as $||{\bf{X}}||_{max} = max_{i,j}|X_{ij}|$
These norms satisfy the inequalities $||{\bf{X}}||_{2} \leq ||{\bf{X}}||_{F} \leq \sqrt{P} ||{\bf{X}}||_{2}$ and $||{\bf{X}}||_{max} \leq ||{\bf{X}}||_{2} \leq \sqrt{MK \times NK} ||{\bf{X}}||_{max}$. In the remainder of our analysis, all inequalities use the matrix 2 norms unless mentioned otherwise. The reader can interpret our results in terms of other norms if needed.


\subsubsection{Perturbation Theory}

How are the singular values and singular vectors of 

\be 
\hat{\bf{{X}}} = {\bf{X}} + {\bf{E}} = \sum\limits_{i=1}^{i=P} \hat{s_{i}}\hat{u_{i}}\hat{v_{i}}^{T}, \enum{svd2}
\ee

related to the singular values and vectors of ${\bf{X}}$? To derive intuition into this, we generate an adversarial image corresponding to figure \ref{fig:fig_2a}, and plot the relative change in the singular values and vectors of the clean matrix (figure \ref{fig:fig_3}). From the figure, we note that the relative change in the small singular values and corresponding singular vectors are large. This intuition can be formalized using results from perturbation theory. For singular values, we have the  results from Weyl  and Mirsky (see \cite{stewart1998perturbation} for details), which state that $|s_{i} - \hat{s_{i}}| \leq ||{\bf{E}}||_{2}$, and  $\sqrt{\sum\limits_{i=1}^{i=P}(s_{i} - \hat{s_{i}})^{2}} \leq ||{\bf{E}}||_{F}$ respectively.
These results hold for all values of ${\bf{E}}$, and show that the absolute change in $s_{i}$ is bounded by the magnitude of the perturbation. When ${\bf{E}}$ is small, we have the more precise result

\be 
\hat{s_{i}} \approx s_{i} + v_{i}^T{\bf{E}}u_{i} + O(||{\bf{E}}||^{2}). \enum{p3}
\ee 

Equation \eref{p3} show that when $s_{i}$ is small, the relative change in its value will be large, and when $s_{i}$ is large, the relative change in its value will be small. 
For singular vectors, a slight re-formulation of Wedin's theorem \cite{o2017random} states that 

\be 
\mbox{Max}(sin(\angle u_{i},\hat{u_{i}}),sin(\angle v_{i},\hat{v_{i}}))\leq \frac{2||{\bf{E}}||}{max(|s_{i}-s_{i+1}|,|s_{i}-s_{i-1}|)}, \enum{wed1}
\ee 
Equation \eref{wed1} shows that the more `spread out' the singular values, the smaller the relative change in the singular vectors. Note from figure \ref{fig:fig_2b}, that the larger singular values are more spread out than the smaller singular values; this finding seems to generally hold true for images. Therefore, in general, we expect the relative change in the singular vectors corresponding to the large singular values will be small and vice-versa.

\subsection{A test for identifying adversarial images}



Assuming that the adversarial perturbation satisfies $||{\bf{E}}|| \geq \alpha ||{\bf{X}}||$ for some small values of $\alpha$, we now derive a metric for separating adversarial images from clean ones.

\subsubsection{The idea behind the test}

Equation \eref{p3} shows that when the magnitude of the singular value is the same order of magnitude as $||{\bf{E}}||$, the change in the singular value will be of the same order of magnitude as the singular value itself. 

Thus, if $s_{m} \leq \alpha ||{\bf{X}}|| \leq ||{\bf{E}}||$, we expect $s_{i}$ to be dramatically different from $\hat{s_{i}}$ whenever $ i \geq m $. Accordingly, we expect the distribution of 
\be
\rho = \sum \limits_{i=m}^{i=P} s_{i}^{2} \enum{rho}
\ee to be dramatically different for the adversarial and clean images. We use this as a basis for separating clean images from adversarial ones, as described below.




\subsubsection{The test}
Given a train dataset, $D_{train}$  of clean images, our test proceeds as follows:

\begin{enumerate}
\item Set $\alpha = 0.01$, and choose $m$ to be the smallest integer which satisfies $s_{i} \leq \alpha s_{1}$ for $95\%$ of the images in $D_{train}$
\item Compute the empirical distribution of $\rho = \sum \limits_{i=m}^{i=P} s_{i}^{2}$ using $D_{train}$, and identify two points, $L$ and $U$, corresponding to the $5^{th}$ and $95^{th}$ percentile of this distribution respectively.
\item Mark a new image as `clean' if the value of $\rho$ for that image lies between $L$ and $U$, and mark it is adversarial otherwise. 
\end{enumerate}
Our test has a false positive rate of $10\%$ i.e., $10\%$ of the clean images will be marked as adversarial. This false positive rate can be regulated by adjusting the values of $L$ and $U$ to meet the modeler's convenience. 








\begin{table}
\scalebox{0.9}
{
\begin{tabular}{ |p{1cm}||p{1.3cm}|p{1.3cm}|p{1.9cm}|  }
 \hline
 \multicolumn{4}{|c|}{MNIST} \\
 \hline
${\bf{\epsilon}}$ & {\bf{FGSM}} &  {\bf{Madry}} & {\bf{Momentum}}\\
 \hline
0.01   & 100\%    &100\% &   100\%\\
 0.03&   100\%  & 100\%  & 100\%\\
0.1 & 100\% & 100\% &  100\%\\
 0.3    &100\% & 100\%&  100\%\\
 \hline
\end{tabular}
\quad 
\begin{tabular}{ |p{1cm}||p{1.3cm}|p{1.3cm}|p{1.9cm}|  }
 \hline
 \multicolumn{4}{|c|}{CIFAR 10} \\
 \hline
${\bf{\epsilon}}$ & {\bf{FGSM}} &  {\bf{Madry}} & {\bf{Momentum}}\\
 \hline
0.01   & 90\%    &90\% &   90\%\\
 0.03&   90\%  & 90\%  & 90\%\\
0.1 & 89\% & 89\% &  88\%\\
 0.3    &81\% & 79\%&  78\%\\
 \hline
\end{tabular}
}
\bigskip
\caption{\label{table1}Percentage of adversarial images that are correctly identified by our metric, when the MNIST (left) and CIFAR 10 (right) datasets are attacked by different strategies and different perturbation strengths. }

\end{table}



\section{Experiments}



In this section, we verify the efficacy of our approach using different datasets and different attack methods.

\subsection{The datasets}
For our experiments, we use the MNIST and CIFAR 10 datasets, which are known to behave very differently in their response to adversarial perturbations \cite{carlini2017adversarial}.

{\bf{The MNIST dataset}} \cite{lecun1998mnist} comprises of 70,000, $28 \times 28$ grayscale images of handwritten digits between 0 and 9. The dataset is split into 60,000 train images and 10,000 validation images. A simple convolutional neural network 
\footnote{Network architecture can be found at \url{https://github.com/keras-team/keras/blob/master/examples/mnist_cnn.py}} achieves an accuracy of $99.25\%$ on the validation dataset; we refer to this network as $M1$.

{\bf{The CIFAR 10 dataset}} \cite{krizhevsky2009learning} comprises 60,000, $32 \times 32$ RBG images of ten objects (car, airplane etc.). The dataset is split into 50,000 train images and 10,000 validation images. A neural network based on the VGG architecture \footnote{Network architecture can be found at \url{https://github.com/geifmany/cifar-vgg}} detailed in \cite{geifman2017selective} achieves an accuracy of $93.56\%$ on the validation dataset; we refer to this network as $M2$.






\subsection{The attack methods}

For our experiments, we generate adversarial images using the Fast Gradient Sign Attack, the Madry attack, and the Momentum attack.

{\bf{The Fast Gradient Sign method}} \cite{kurakin2016adversarial} is an iterative attack which generates adversarial images by perturbing clean image pixels in the direction corresponding to the sign of the loss function. This is one of the first known methods used for generating adversarial images. 

{\bf{The Madry attack}} \cite{madry2017towards} is an optimization based attack which uses Projected Gradient Descent to generate the strongest `first order attack' using local information in the neural network.  \footnote{This method is chosen over the equally competent optimization based attack by Carlini and Wagner \cite{carlini2017towards} because it appears to be marginally better at generating adversarial images.}

{\bf{The Momentum attack}} \cite{dong2017boosting} is a momentum iterative gradient based attack which boosts the success rates of adversarial images.  This attack won the first places in NIPS 2017 `Non-targeted Adversarial Attacks' and `Targeted Adversarial Attacks' challenges.






These attacks are chosen because 1) they are state of the art and 2) because they use extremely different approaches to generate adversarial images. Each of these attacks has a parameter, $\epsilon$, which describes the magnitude of the maximum difference 
between a pixel in the pre-processed clean image and a pixel in the pre-processed adversarial image; the larger the value of $\epsilon$, the more effective the attack. In our experiments, we use the Cleverhans toolbox \cite{papernot2017cleverhans} to simulate each of these attacks

\subsection{The experiment}

For our experiments, we first compute the values of $L$ and $U$ using the train dataset. Next, we attack the test dataset using different attack strategies and perturbation strengths, and evaluate the percentage of the adversarial images that are correctly classified by our approach. As a control, we also evaluate  the percentage of adversarial images identified in the clean test dataset. 

For the MNIST train dataset, the values of $L$ and $U$ are $8.27 \times 10^{-16}$ and $0.0027$ respectively. With these values of $L$ and $U$, approximately $10\%$ of the images in the clean MNIST test dataset are identified as adversarial; this is consistent with our false positive rate of $10 \%$. From table \ref{table1}, we note that our test is able to correctly identify all of the adversarial images that are obtained by attacking $M1$ using different attack strategies, and different values of $\epsilon \geq 0.01$.

For the CIFAR 10 train dataset, the values of $L$ and $U$ are $3.65 \times 10^{-5}$ and $0.008$ respectively. Here too, the false positive rate on the clean test dataset is approximately 10\%. From table \ref{table1}, we note that although our test is able to correctly identify most of the adversarial images obtained by attacking $M2$, the accuracy is lower than that with the MNIST dataset. This finding is consistent with the observation of \cite{carlini2017adversarial}, that adversarial perturbations are harder to detect in the CIFAR 10 dataset than in the MNIST dataset.

\subsection{The problems with extremely small perturbations}

Our analysis assumes that the strength of the adversarial perturbation is bounded below by a value comparable to the smaller singular values of the image matrix. When the magnitude of the adversarial perturbation is extremely small, this assumption may not hold true. Indeed, our approach is unable to identify adversarial images generated using the Fast Gradient Sign Attack with $\epsilon = 0.0001$. Similar observations have been made in \cite{grosse2017statistical} and \cite{gong2017adversarial}.

Equations \eref{p3} and \eref{rho} suggest that this is to be expected; the difference between the value of $\rho$ for the clean and adversarial image matrix will be $O(\epsilon^{2})$, and when $\epsilon$ is extremely small, the change in the distribution of $\rho$ for the clean image matrices will not be statistically significant.



\section{General insights into adversarial perturbations}

In this section, we use our approach to provide general geometric insights into several properties of adversarial perturbations.

\subsection{Why don't image rotations nullify adversarial perturbations?}

We know that rotating an image does not nullify the effects of adversarial perturbations \cite{athalye2017synthesizing}, but the reason for this is still largely unknown. To derive insights into why this might be the case, we consider a general rotation, ${\bf{Y}} = {\bf{P}}{\bf{X}}{\bf{Q}}^{T} $ of the clean image matrix, where
${\bf{P}}$ and ${\bf{Q}}$ are orthonormal matrices determining the angle of rotation. Since rotating an image with $K$ channels involves rotating each of the channels by the same angle, these orthonormal matrices will take the form
\[
  {\bf{P}} =
  \begin{bmatrix}
    {\bf{P}}_{1} & 0 & 0 & \dots  & 0\\
    0 & {\bf{P}}_{1} & 0 & \dots  & 0\\
    \vdots & \vdots & \ddots & \vdots & \vdots \\
    0  & \dots & 0 & {\bf{P}}_{1} & 0\\
    0 & 0  & \dots & 0  & {\bf{P}}_{1}
\end{bmatrix}
\quad\mbox{  and  } \quad
  {\bf{Q}} =
  \begin{bmatrix}
    {\bf{Q}}_{1} & 0 & 0 & \dots  & 0\\
    0 & {\bf{Q}}_{1} & 0 & \dots  & 0\\
    \vdots & \vdots & \ddots & \vdots & \vdots \\
    0  & \dots & 0 & {\bf{Q}}_{1} & 0\\
    0 & 0  & \dots & 0  & {\bf{Q}}_{1}
\end{bmatrix},
\]
where ${\bf{P}}_{1}$ and ${\bf{Q}}_{1}$ are orthonormal rotation matrices describing the rotation of each individual channel.

If the SVD of the clean image matrix, ${\bf{X}}$, is given by \eref{svd1}, then the SVD of the rotated clean image matrix, ${\bf{Y}}$ will be given by ${\bf{Y}} = {\bf{P}}{\bf{X}}{\bf{Q}}^{T} =  \sum s_{i}p_{i}q_{i}^{T}$, where  $p_{i} = {\bf{P}}u_{i}$, and $q_{i} = {\bf{Q}}v_{i}$ is the $i^{th}$ left and right singular vectors of ${\bf{Y}}$ respectively.
Similarly, if the SVD of the adversarial image matrix, ${\bf{\hat{X}}}$ is given by \eref{svd2}, then the SVD of the rotated adversarial image matrix, $\hat{{\bf{Y}}}$, will be given by $\hat{{\bf{Y}}} = {\bf{P}}{\hat{\bf{X}}}{\bf{Q}}^{T} = \sum \hat{s_{i}}\hat{p_{i}}\hat{q_{i}}^{T},$ where $\hat{p_{i}} = {\bf{P}}\hat{u_{i}}$ and $\hat{q_{i}} = {\bf{Q}}\hat{v_{i}}$ are the $i^{th}$ left and right singular vectors of ${\bf{\hat{Y}}}$ respectively.


Note that rotating an image does not alter the singular values of the image matrix. Furthermore, since ${\bf{P}}$ and ${\bf{Q}}$ are orthornomal, we have $p_{i}^{T}\hat{p_{i}}  = ({\bf{P}}u_{i})^{T}({\bf{P}}\hat{u_{i}}) = u_{i}^{T}\hat{u_{i}}$, and $q_{i}^{T}\hat{q_{i}} = ({\bf{Q}}v_{i})^{T}({\bf{Q}}\hat{v_{i}}) = v_{i}^{T}\hat{v_{i}}$. Therefore, the relative change in the singular values and vectors 
of the clean image matrix on applying an adversarial perturbation are not effected by rotating the frame of reference. Our approach relies on only the singular values of the image matrix and therefore,  if $\hat{{\bf{X}}}$ is marked as adversarial, then $\hat{{\bf{Y}}} = {\bf{P}}\hat{{\bf{X}}}{\bf{Q}}^{T}$, which has the same singular values as $\hat{{\bf{X}}}$  will also be marked as adversarial.

\subsection{Why does image compression reverse small adversarial perturbations?}

In \cite{dziugaite2016study}, the authors show that image compression reverses small adversarial perturbation. As stated in that paper, the reason for this is not known. Here, we provide a possible explanation for these findings. We begin by considering images with one channel. With the SVD of the clean image matrix, ${\bf{X}}$ given by \eref{svd1}, we define an `extremely small' adversarial perturbation as follows:

\begin{defn} An adversarial perturbation, ${\bf{E}}$ is said to be `extremely small' if ${\hat{\bf{X}}} = {\bf{X}} + {\bf{E}} $ describes an adversarial image matrix, and there exists a value of $k$ such that

\begin{enumerate}
\item $||{\bf{E}}||_{2} \ll s_{i} $ for all $i \leq k$,
\item $||{\bf{E}}||_{2} \ll (s_{i} - s_{i+1}) $  for all $i \leq k$ and,
\item ${\bf{Y}} = \sum\limits_{i=1}^{i=k} s_{i}u_{i}v_{i}^{T}$ has the same label as ${\bf{X}}$.
\end{enumerate}
\end{defn}
\noindent

In the above definition, the first 
two requirements state that the size of the perturbation is smaller than the magnitude of the first $k$ singular values, and the separation between them. These requirements are motivated by the facts that 1) the top few singular values of the image matrix are typically larger and more spread out than the remaining singular values (see figure \ref{fig:fig_2b}), and 2) that the adversarial perturbation is `extremely small', and is therefore comparable to one of the smallest singular value of ${\bf{X}}$. The third requirement can be thought of as a consequence of the first two. Since most of the information about the image is contained in the top $k$ singular values, we expect that the neural network will assign the same label to  ${\bf{X}}$ and ${\bf{Y}}$. The third requirement simply guarantees this fact.

From defenition 1, and equations \eref{p3} and \eref{wed1}, we have $s_{i} \approx \hat{s_{i}}$, $u_{i} \approx \hat{u_{i}}$, and $v_{i} \approx \hat{v_{i}}$ for all $i \leq k$. This implies that SVD of ${\bf{\hat{X}}}$ can be written as 

\be 
{\bf{\hat{X}}} \approx \sum\limits_{i=1}^{i=k} s_{i}u_{i}v_{i}^{T} + \sum\limits_{i=k+1}^{i=P}\hat{s_{i}}\hat{u_{i}}\hat{v_{i}}^{T} =  {\bf{Y}} + \sum\limits_{i=k+1}^{i=P}\hat{s_{i}}\hat{u_{i}}\hat{v_{i}}^{T}
\ee 

Since ${\bf{X}}$ and ${\bf{Y}}$ have the same label, the adversarial perturbation can be reversed by compressing the adversarial image matrix i.e., by 
setting $\hat{s}_{k+1}, \hat{s}_{k+2} \hdots \hat{s}_{P} =0$. For a single channel image, compressing the adversarial image matrix is equivalent to compressing the adversarial image and therefore, compressing an adversarial image can reverse extremely small perturbations. The result can be extended to the case of multi-channel images by repeating the analysis for each channel in the image. 

\subsection{Why are adversarial perturbation imperceptible to humans?}

The large singular values and corresponding singular vectors contain most of the information about the image and likely play the most important role in human perception. Indeed, this is the basis for image compression described above. Our analysis suggests that adversarial perturbations will have little effect on these large singular values and singular vectors and therefore, we expect clean and adversarial images to be perceptually similar.



\section{Conclusion}

In this paper, we derive a metric which 
identifies adversarial images across a variety of datasets and attack strategies. The advantage of our approach over previous studies is that it does not have to be re-calibrated every time the method for generating adversarial images is changed. Our approach provides geometric insights into 1) why image rotations do not counter adversarial perturbations, and 2) why image compressions can counter small adversarial perturbations. In future work, we plan to use our results to develop a defense which counters the effects of adversarial perturbations.

\bibliographystyle{unsrt}

\begin{thebibliography}{10}

\bibitem{moosavi2016deepfool}
Seyed~Mohsen Moosavi~Dezfooli, Alhussein Fawzi, and Pascal Frossard.
\newblock Deepfool: a simple and accurate method to fool deep neural networks.
\newblock In {\em Proceedings of 2016 IEEE Conference on Computer Vision and
  Pattern Recognition (CVPR)}, number EPFL-CONF-218057, 2016.

\bibitem{goodfellow2014explaining}
Ian~J Goodfellow, Jonathon Shlens, and Christian Szegedy.
\newblock Explaining and harnessing adversarial examples.
\newblock {\em arXiv preprint arXiv:1412.6572}, 2014.

\bibitem{madry2017towards}
Aleksander Madry, Aleksandar Makelov, Ludwig Schmidt, Dimitris Tsipras, and
  Adrian Vladu.
\newblock Towards deep learning models resistant to adversarial attacks.
\newblock {\em arXiv preprint arXiv:1706.06083}, 2017.

\bibitem{mcdaniel2016machine}
Patrick McDaniel, Nicolas Papernot, and Z~Berkay Celik.
\newblock Machine learning in adversarial settings.
\newblock {\em IEEE Security \& Privacy}, 14(3):68--72, 2016.

\bibitem{evtimov2017robust}
Ivan Evtimov, Kevin Eykholt, Earlence Fernandes, Tadayoshi Kohno, Bo~Li, Atul
  Prakash, Amir Rahmati, and Dawn Song.
\newblock Robust physical-world attacks on machine learning models.
\newblock {\em arXiv preprint arXiv:1707.08945}, 2017.

\bibitem{grosse2017statistical}
Kathrin Grosse, Praveen Manoharan, Nicolas Papernot, Michael Backes, and
  Patrick McDaniel.
\newblock On the (statistical) detection of adversarial examples.
\newblock {\em arXiv preprint arXiv:1702.06280}, 2017.

\bibitem{gong2017adversarial}
Zhitao Gong, Wenlu Wang, and Wei-Shinn Ku.
\newblock Adversarial and clean data are not twins.
\newblock {\em arXiv preprint arXiv:1704.04960}, 2017.

\bibitem{dziugaite2016study}
Gintare~Karolina Dziugaite, Zoubin Ghahramani, and Daniel~M Roy.
\newblock A study of the effect of jpg compression on adversarial images.
\newblock {\em arXiv preprint arXiv:1608.00853}, 2016.

\bibitem{szegedy2013intriguing}
Christian Szegedy, Wojciech Zaremba, Ilya Sutskever, Joan Bruna, Dumitru Erhan,
  Ian Goodfellow, and Rob Fergus.
\newblock Intriguing properties of neural networks.
\newblock {\em arXiv preprint arXiv:1312.6199}, 2013.

\bibitem{carlini2017towards}
Nicholas Carlini and David Wagner.
\newblock Towards evaluating the robustness of neural networks.
\newblock In {\em Security and Privacy (SP), 2017 IEEE Symposium on}, pages
  39--57. IEEE, 2017.

\bibitem{akhtar2018threat}
Naveed Akhtar and Ajmal Mian.
\newblock Threat of adversarial attacks on deep learning in computer vision: A
  survey.
\newblock {\em arXiv preprint arXiv:1801.00553}, 2018.

\bibitem{kurakin2016adversarial}
Alexey Kurakin, Ian Goodfellow, and Samy Bengio.
\newblock Adversarial examples in the physical world.
\newblock {\em arXiv preprint arXiv:1607.02533}, 2016.

\bibitem{papernot2016distillation}
Nicolas Papernot, Patrick McDaniel, Xi~Wu, Somesh Jha, and Ananthram Swami.
\newblock Distillation as a defense to adversarial perturbations against deep
  neural networks.
\newblock In {\em Security and Privacy (SP), 2016 IEEE Symposium on}, pages
  582--597. IEEE, 2016.

\bibitem{guo2017countering}
Chuan Guo, Mayank Rana, Moustapha Ciss{\'e}, and Laurens van~der Maaten.
\newblock Countering adversarial images using input transformations.
\newblock {\em arXiv preprint arXiv:1711.00117}, 2017.

\bibitem{carlini2017adversarial}
Nicholas Carlini and David Wagner.
\newblock Adversarial examples are not easily detected: Bypassing ten detection
  methods.
\newblock In {\em Proceedings of the 10th ACM Workshop on Artificial
  Intelligence and Security}, pages 3--14. ACM, 2017.

\bibitem{tramer2017space}
Florian Tram{\`e}r, Nicolas Papernot, Ian Goodfellow, Dan Boneh, and Patrick
  McDaniel.
\newblock The space of transferable adversarial examples.
\newblock {\em arXiv preprint arXiv:1704.03453}, 2017.

\bibitem{elsayed2018adversarial}
Gamaleldin~F Elsayed, Shreya Shankar, Brian Cheung, Nicolas Papernot, Alex
  Kurakin, Ian Goodfellow, and Jascha Sohl-Dickstein.
\newblock Adversarial examples that fool both human and computer vision.
\newblock {\em arXiv preprint arXiv:1802.08195}, 2018.

\bibitem{su2017one}
Jiawei Su, Danilo~Vasconcellos Vargas, and Sakurai Kouichi.
\newblock One pixel attack for fooling deep neural networks.
\newblock {\em arXiv preprint arXiv:1710.08864}, 2017.

\bibitem{horn1990matrix}
Roger~A Horn, Roger~A Horn, and Charles~R Johnson.
\newblock {\em Matrix analysis}.
\newblock Cambridge university press, 1990.

\bibitem{kato2013perturbation}
Tosio Kato.
\newblock {\em Perturbation theory for linear operators}, volume 132.
\newblock Springer Science \& Business Media, 2013.

\bibitem{stewart1998perturbation}
Gilbert~W Stewart.
\newblock Perturbation theory for the singular value decomposition.
\newblock Technical report, 1998.

\bibitem{o2017random}
Sean O'Rourke, Van Vu, and Ke~Wang.
\newblock Random perturbation of low rank matrices: Improving classical bounds.
\newblock {\em Linear Algebra and its Applications}, 2017.

\bibitem{lecun1998mnist}
Yann LeCun.
\newblock The mnist database of handwritten digits.
\newblock {\em http://yann. lecun. com/exdb/mnist/}, 1998.

\bibitem{krizhevsky2009learning}
Alex Krizhevsky and Geoffrey Hinton.
\newblock Learning multiple layers of features from tiny images.
\newblock 2009.

\bibitem{geifman2017selective}
Yonatan Geifman and Ran El-Yaniv.
\newblock Selective classification for deep neural networks.
\newblock In {\em Advances in neural information processing systems}, pages
  4885--4894, 2017.

\bibitem{dong2017boosting}
Yinpeng Dong, Fangzhou Liao, Tianyu Pang, Hang Su, Xiaolin Hu, Jianguo Li, and
  Jun Zhu.
\newblock Boosting adversarial attacks with momentum. arxiv preprint.
\newblock {\em arXiv preprint arXiv:1710.06081}, 2017.

\bibitem{papernot2017cleverhans}
Ian Goodfellow Reuben Feinman Fartash Faghri Alexander Matyasko Karen
  Hambardzumyan Yi-Lin Juang Alexey Kurakin Ryan Sheatsley Abhibhav Garg
  Yen-Chen~Lin Nicolas~Papernot, Nicholas~Carlini.
\newblock cleverhans v2.0.0: an adversarial machine learning library.
\newblock {\em arXiv preprint arXiv:1610.00768}, 2017.

\bibitem{athalye2017synthesizing}
Anish Athalye and Ilya Sutskever.
\newblock Synthesizing robust adversarial examples.
\newblock {\em arXiv preprint arXiv:1707.07397}, 2017.

\end{thebibliography}

\end{document}